\documentclass[letterpaper, 10 pt, conference]{ieeeconf}  
\IEEEoverridecommandlockouts                              
\usepackage{url}
\usepackage{graphicx}
\usepackage{hyperref}
\usepackage{listings}
\usepackage{xcolor,colortbl}
\title{\LARGE \bf
OpenDR: An Open Toolkit for Enabling High Performance, Low Footprint Deep Learning for Robotics
}

\author{N. Passalis$^{1}$, S. Pedrazzi$^{2}$, R. Babuska$^{3}$, W. Burgard$^{4}$, 
D. Dias$^{2}$, F. Ferro$^{5}$, M. Gabbouj$^{6}$, O. Green$^{7}$, 
A. Iosifidis$^{8}$, \\ E. Kayacan$^{8}$,  J. Kober$^{3}$, O. Michel$^{2}$, N. Nikolaidis$^{1}$, P. Nousi$^{1}$, R. Pieters$^{6}$, M. Tzelepi$^{1}$,  A. Valada$^{4}$, and A. Tefas$^{1}$
\thanks{$^{1}$N. Passalis, N. Nikolaidis, P. Nousi, M. Tzelepi, and A. Tefas are with Dept. of Informatics, Aristotle University of Thessaloniki, Greece. {\tt\footnotesize \{passalis, nnik, paranous, tzelepi, tefas\}@csd.auth.gr}}
\thanks{$^{2}$S. Pedrazzi, D. Dias and O. Michel are with Cyberbotics, Switzerland. {\tt\footnotesize \{stefania.pedrazzi, daniel.dias, olivier.michel\}@cyberbotics.com}}%
\thanks{$^{3}$R. Babuska and J. Kober are with Dept. of Cognitive Robotics, Delft University of Technology, The Netherlands. {\tt\footnotesize \{r.babuska, j.kober\}@tudelft.nl	}}%
\thanks{$^{4}$W. Burgard and A. Valada are with Dept. of Computer Science, University of Freiburg, Germany. {\tt\footnotesize \{burgard, valada\}@cs.uni-freiburg.de	}}%
\thanks{$^{5}$F. Ferro is with PAL Robotics, Spain. {\tt\footnotesize francesco.ferro@pal-robotics.com }}%
\thanks{$^{6}$M. Gabbouj and R. Pieters are with the units of Computing Sciences, and Automation Technology and Mechanical Engineering, Tampere University, Finland. {\tt\footnotesize \{moncef.gabbouj, roel.pieters\}@tuni.fi	}}%
\thanks{$^{7}$O. Green is with Agrointelli, Denmark. {\tt\footnotesize olg@agrointelli.com}}%
\thanks{$^{8}$ A. Iosifidis and E. Kayacan are with the Department of Electrical and Computer Engineering, Aarhus University, Denmark. {\tt\footnotesize \{ai, erdal\}@ece.au.dk}}%
\thanks{This work was supported by the European Union's Horizon 2020 Research and Innovation Program (OpenDR) under Grant 871449. This publication reflects the authors' views only. The European Commission is not responsible for any use that may be made of the information it contains.}        
}

\begin{document}

\maketitle
\thispagestyle{empty}
\pagestyle{empty}

\begin{abstract}
Existing Deep Learning (DL) frameworks typically do not provide ready-to-use solutions for robotics, where very specific learning, reasoning, and embodiment problems exist. Their relatively steep learning curve and the different methodologies employed by DL compared to traditional approaches, along with the high complexity of DL models, which often leads to the need of employing specialized hardware accelerators,  further increase the effort and cost needed to employ DL models in robotics. Also, most of the existing DL methods follow a static inference paradigm, as inherited
by the traditional computer vision pipelines, ignoring active perception, which can be employed to actively interact with the environment in order to increase perception accuracy. In this paper, we present the Open Deep Learning Toolkit for Robotics (OpenDR). OpenDR aims at developing an open, non-proprietary, efficient, and  modular toolkit that can be easily used by robotics companies and research institutions to efficiently develop and deploy AI and cognition technologies to robotics applications, providing a solid step towards addressing the aforementioned challenges. We also detail the design choices, along with an abstract interface that was created to overcome these challenges. This interface can describe various robotic tasks, spanning beyond traditional DL cognition and inference, as known by existing frameworks, incorporating openness, homogeneity and robotics-oriented perception e.g., through active perception, as its core design principles. 
\end{abstract}

\begin{figure}
    \centering
    \includegraphics[width=0.59\linewidth]{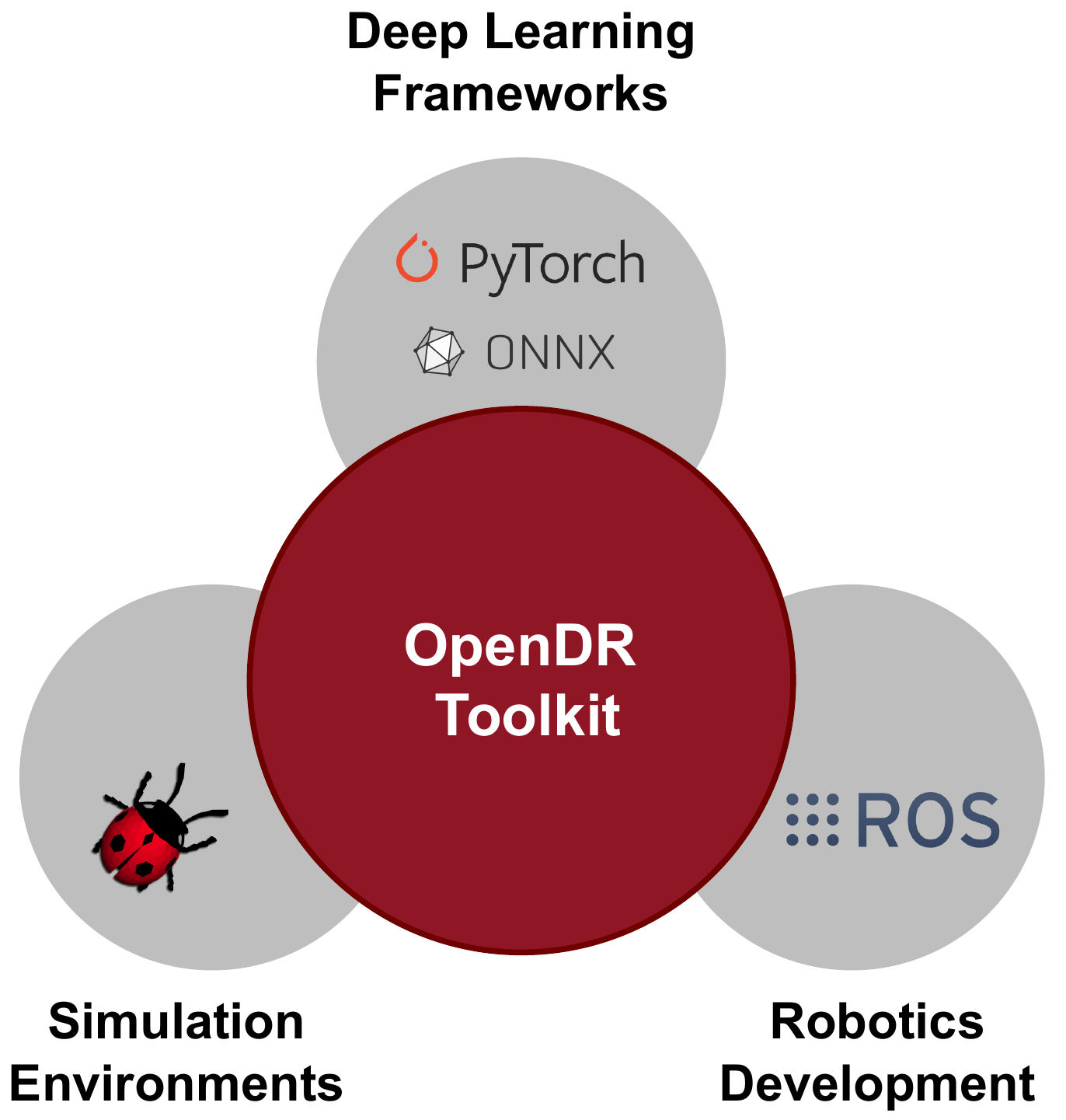}
    \caption{OpenDR lies at the intersection between deep learning frameworks, simulation environments and robotics development tools}
    \label{fig:opendr}
\end{figure}

\section{INTRODUCTION}

Deep Learning (DL) led to a number of spectacular applications, ranging from self-driving cars to robots that outperform humans in various tasks~\cite{lecun2015deep}. However, employing DL in robotics leads to very specific learning, reasoning, and embodiment problems and research questions that are typically not addressed by the computer vision and machine learning communities. At the same time, existing DL frameworks, such as PyTorch~\cite{paszke2019pytorch}, usually do not provide ready-to-use solutions for robotics, have a steep learning curve and employ a different methodology than traditional approaches. Furthermore, the high complexity of DL models often leads to the need of employing specialized accelerators, in order to ensure that models will be able to be successfully deployed in edge devices with limited computational power, which further increases the cost and effort required to incorporate DL models into robotic systems. At the same time, despite these recent achievements of DL, most of the existing methods also suffer from a significant
drawback: they follow a static inference paradigm, as inherited
by the traditional computer vision pipelines. More specifically,
DL models perform inference on a fixed and static input, ignoring that robots have the ability to interact with the environment to better sense their surroundings. This approach, usually referred to as called active perception~\cite{bajcsy2018revisiting}, allows for manipulating
the robot/sensor to acquire a better and more clean
view/signal.

Therefore, the need for an open DL toolkit that contains easy to train and deploy real-time, lightweight, and efficient DL models for robotics is evident. In this paper, we present the Open Deep Learning Toolkit for Robotics (OpenDR). OpenDR aims at developing an open, non-proprietary modular toolkit that can be easily used by robotics companies and research institutions to efficiently develop and deploy AI and cognition technologies to robotics applications. At a high level, OpenDR contains a selection of cognition, perception and robot action and decision making algorithms, along with general-purpose functionalities that are necessary for common robotics tasks. OpenDR provides an intuitive and easy to use Python interface, a C API for selected tools, a wealth of usage examples and supporting tools, as well as ready-to-use Robot Operating System (ROS) nodes for various perception tasks. OpenDR is built to support Webots~\cite{michel2004cyberbotics}, while it also extensively follows industry standards, such as ONNX model format~\cite{onnxruntime} and OpenAI Gym Interface~\cite{brockman2016openai}, lying at the intersection between DL frameworks, simulation environments and robotics development tools, as shown in Fig.~\ref{fig:opendr}.

The main contribution of this paper is to present the design principles of OpenDR, as well as the architecture that was developed following these principles. Designing such a toolkit, that will incorporate a wide range of cognition and perception approaches, in a modular and homogeneous manner bears significant challenges, as further explained in this paper. To this end, in this paper we detail the design choices, along with the abstract interface that was created to overcome these challenges. This interface  can describe various robotic tasks, spanning beyond traditional DL cognition and inference, as known by existing frameworks, incorporating openness, homogeneity and robotics-oriented perception e.g., through active perception, as its core design principles. 
The toolkit is available for download at \url{github.com/opendr-eu/opendr} and distributed through multiple channels.

The rest of this paper is structured as follows. In Section~\ref{section:requirements} we introduce the requirements and design principles of OpenDR. Then, in Section~\ref{sec:arch} we describe the architecture of OpenDR, while in Section~\ref{section:engine} we present the \textit{engine} package of OpenDR, which acts as the base for ensuring the homogeneity of the toolkit. Finally, Section~\ref{sec:examples} provides a few usage examples, demonstrating the ease of use of the toolkit, while Section~\ref{sec:conclusions} concludes this paper.

\section{OpenDR Requirements and Design Principles}
\label{section:requirements}
\begin{figure}
    \centering
    \includegraphics[width=0.9\linewidth]{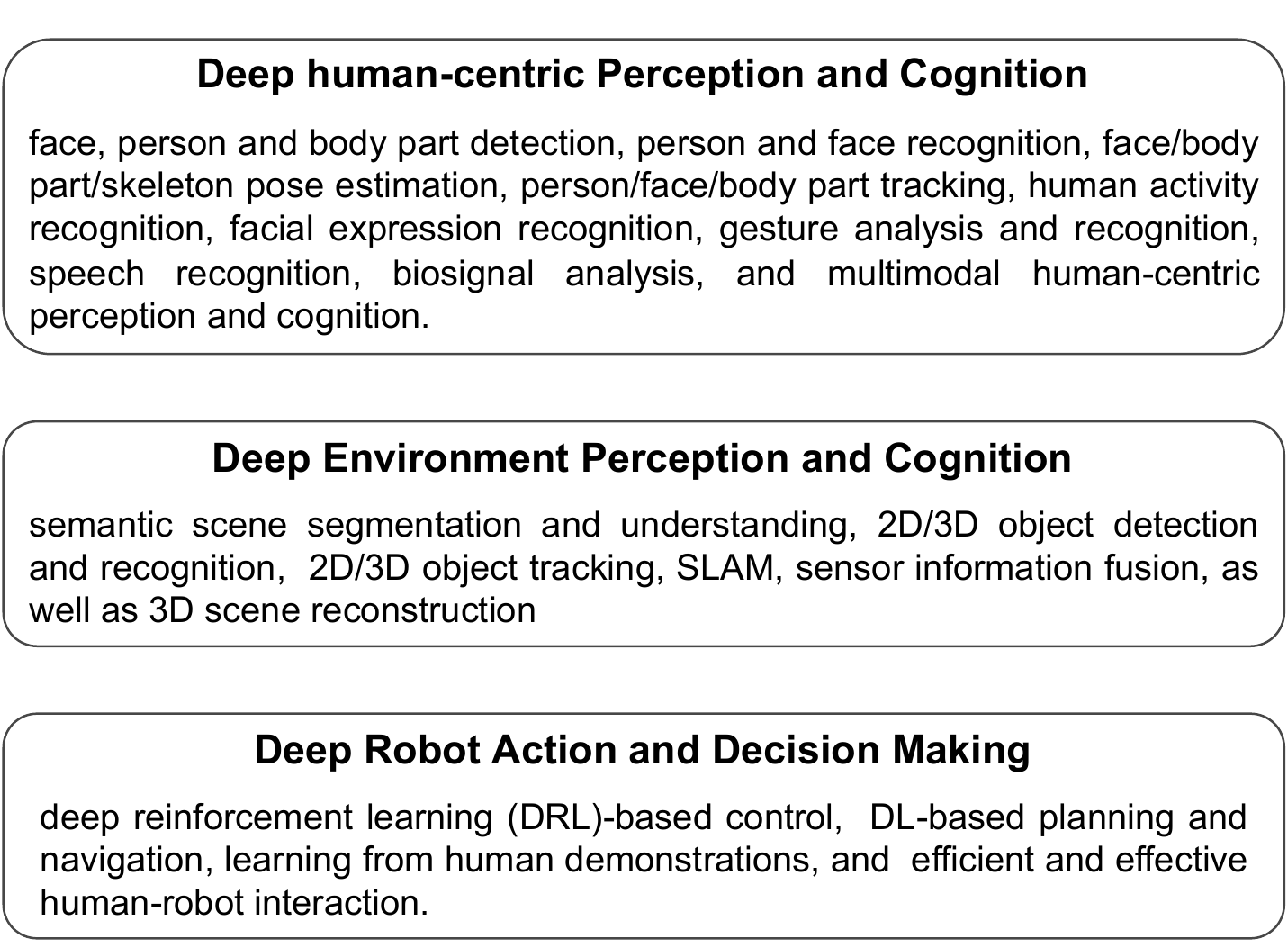}
    \caption{OpenDR aims to cover a wide range of different robotics applications}
    \label{fig:tools}
\end{figure}

This section provides a brief overview of the key requirements and design principles that OpenDR toolkit should meet. OpenDR should have the following system qualities:
\begin{itemize}
    \item \textbf{Ease of use}: OpenDR should be easy to use and well documented so that even new users with minimal background in DL can easily start working with it.
    \item \textbf{Openness}: To maximize the visibility and create a community around the OpenDR toolkit, all toolkit sources should be publicly provided under the Apache license version 2.0.
    \item \textbf{Modularity}: OpenDR should provide a common interface to a set of learning tools and it should be designed in such a way that it is easy to add new functionalities. 
    \item \textbf{Portability}: When technically possible, the toolkit should run on all major operating systems (Linux, Windows, and macOS) and embedded hardware, allowing for easy deployment, as well as easy experimentation in various environments and simulators. CPU execution and (optional) GPU acceleration should be supported.
    \item \textbf{Dependencies}: OpenDR should rely on a consistent and minimal number of third-party libraries, as well as avoid a monolithic design.
    \item \textbf{Interoperability}: OpenDR should be an algorithm and platform agnostic toolkit that could be easily extended to interoperate with widely used applications (e.g., DL libraries, robotics simulators, etc.).
\end{itemize}
At the same time, OpenDR aims to cover a wide range of different applications, including deep human-centric perception and cognition, environment perception and cognition, as well as robot action and decision making, as shown in Fig.~\ref{fig:tools}.
Apart from these basic perception and control functionalities, OpenDR should also provide tools for enabling automatic hyperparameter tuning for Deep Reinforcement Learning (DRL) methods to minimize the effort needed for training such  models.

\begin{figure*}[t!]
    \centering
    \includegraphics[width=0.85\linewidth]{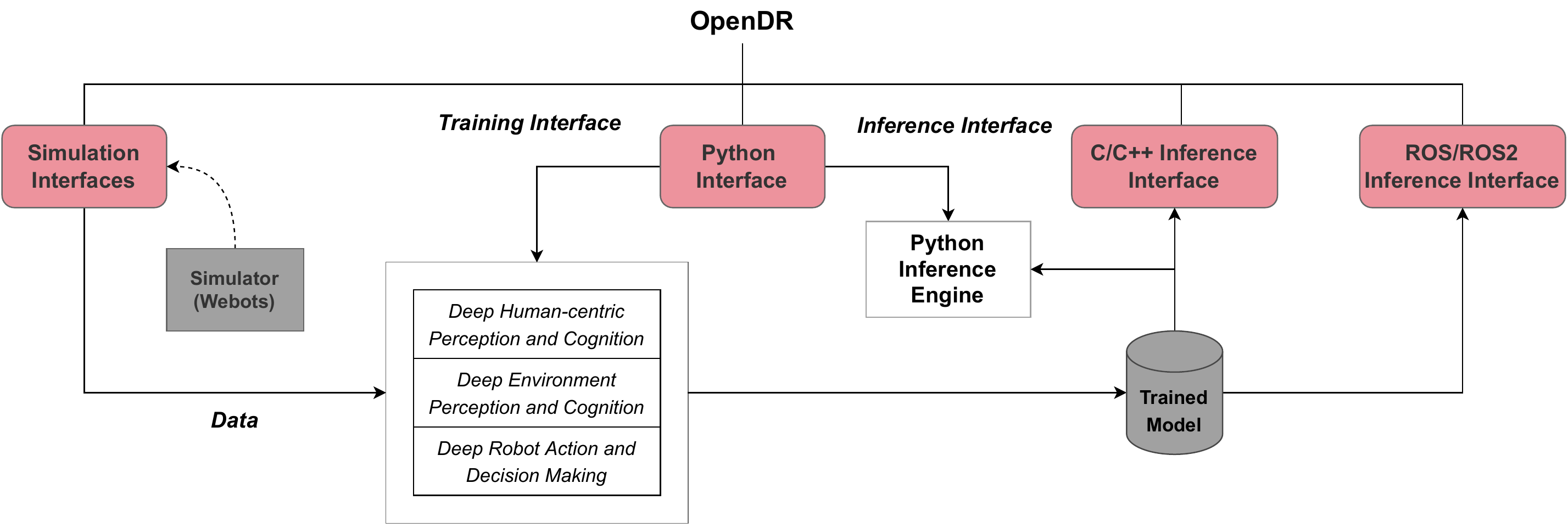}
    \caption{OpenDR toolkit structure}
    \label{fig:opendr-structure}
\end{figure*}

Furthermore, in order to maximize the impact of the developed toolkit in real-world robotics applications, the following requirements regarding DL models have been identified:
\begin{itemize}
    \item \textbf {Pretrained Models:} OpenDR should provide pretrained models to allow users to directly start using the toolkit with minimal time investment. 
    \item \textbf{Active Perception:} OpenDR should provide active perception in a transparent manner, allowing the robotic system to actively interact with its environment in order to improve the perception accuracy.
    \item \textbf{Fine-tuning}: OpenDR should allow pretrained DNN-based controllers to continue learning online, such that it can compensate for internal uncertainties and external disturbances.
    \item \textbf{Prior Knowledge:} OpenDR should allow the user to include prior knowledge on control tasks.
    \item \textbf{Co-integration of simulation and learning:}  OpenDR  should  provide the necessary means to allow for co-integration of simulation and model training, minimizing the effort needed for training and evaluating DL models.
\end{itemize}

OpenDR should also ensure that the provided models are efficient and with realistic requirements, given that many perception tasks are often performed on relatively low-power edge devices. For this reason, the developed methods and models should fulfill some minimal performance requirements.  For each perception task, OpenDR should provide at least one real-time algorithm for a reference embedded GPU-based system. Real-time in OpenDR is defined as achieving 25 Frames Per Second (FPS) when performing inference. At the same time, to ensure that multiple algorithms could be executed on the same device, implementations should aim to use no more than 1GB of RAM per model. Finally, high-resolution analysis in real-time should also be supported for algorithms that can benefit from this, e.g., for detecting very small objects from large distances.

\section{OpenDR ARCHITECTURE}
\label{sec:arch}

OpenDR toolkit has been designed to meet the ease of use, modularity, portability, efficiency, minimal dependency, and interoperability requirements described in Section~\ref{section:requirements}. To this end, the OpenDR toolkit has a modular design and it is composed of three different modules, as summarized in Fig.~\ref{fig:opendr-structure} and described below:
\begin{itemize}
    \item  \textbf{Training Module:} The training module provides a large set of functionalities that allow users to train DL models and export the resulting trained binary model so that it can be used on any supported platform.
    \item  \textbf{Inference Module:} The inference module provides all functionalities required to run the binary trained DL models on the supported platforms. A set of pretrained models have been developed and included in the toolkit to allow users to directly use them for a number of different use cases and applications.
    \item \textbf{Simulation Module:} OpenDR toolkit is developed in such a manner ensuring full compatibility with the Webots simulation software~\cite{michel2004cyberbotics}. Users can model their scenario in Webots and use it to train the DL model or apply pretrained models to complete some tasks and evaluate the performance. The simulation module is responsible for providing the necessary interface to Webots, as well as other data generation and simulation tools.
\end{itemize}
The toolkit core application is written in Python and its structure is depicted in Fig.~\ref{fig:opendr-structure}. Note that the training and inference module are integrated in the Python implementation, since training and inference are closely tied. On the other hand, C/C++ and ROS/ROS2 interfaces for inference and simulation are separated and provided as individual modules. Note that OpenDR extensively relies on cross-platform tools and frameworks, such as Python and PyTorch, ensuring that the toolkit can be used on a wide variety of different platforms. OpenDR also enforces the homogeneity of the toolkit using standardized interfaces through abstract class definitions, as described below.

\noindent  \textbf{Python API} The Python API of OpenDR is split into five discrete packages:
\begin{enumerate}
    \item \textit{engine}, which contains re-usable definitions of various classes that are needed to implement the core functionality of OpenDR,
    \item \textit{perception}, which contains implementations of DL algorithms for various perception tasks,
    \item \textit{control}, which contains implementation of DL-based control algorithms,
    \item \textit{planning}, which contains implementations of DL-based planning algorithms, and
    \item  \textit{simulation}, which contains the necessary tools for interfacing the learning algorithms with simulation environments.
\end{enumerate}
All algorithm implementations are based on the engine package to ensure code uniformity and modularity. The interactions between the various engine-related classes are further discussed in Section~\ref{section:engine}.

\noindent \textbf{C/C++ API} Apart from providing a training and inference API in Python, OpenDR also provides a C/C++ inference API to allow for running the developed models for high-performance applications as efficiently as possible, e.g., on embedded devices. Since this API will be only needed for inference, its function is greatly simplified compared to the Python API. Specifically, for each algorithm supported by the C/C++ API the following two functions are supported: \texttt{load\_X()} and \texttt{infer\_X()}, where \texttt{X} is the name of the implemented algorithm. The first one is responsible for loading the model saved in OpenDR format in the appropriate form for inference, while the second one is for performing inference using the saved model. Additional structures and auxiliary functions for supporting the inference process are defined following the Python data and target class definitions and provided as standalone functions. Similar to the Python API, these definitions are shared among the algorithms to ensure the homogeneity of the toolkit. Note that this API is built leveraging the C++ runtime engine of ONNX, ensuring both the simplicity of the implementation and well high performance with minimal development cost.  Currently, OpenDR provides a C API only for selected tools. 
 
\noindent \textbf{ROS/ROS2 API} For each of the perception, control, and planning tasks, OpenDR also provides standalone ROS nodes, demonstrating the generality and efficiency of the library, while also providing a ready-to-use solution for many different tasks. These nodes act as wrapper nodes to the main Python API. OpenDR re-uses, as much as possible, standard ROS interfaces and messages for the communication between different nodes. Furthermore, OpenDR has developed a bridge subsystem, provided as a standalone ROS package called \texttt{ros\_bridge} which provides an easy to use interface to translate ROS messages into OpenDR data types and vice versa.

\noindent \textbf{Data Handling} A large variety of different formats exist and OpenDR should use a well-defined format to ensure the compatibility of the library with well-established open standards and formats. For example, an image can be read in at least 8 different ways based on the channel ordering (e.g., RGB or BGR), number representation (e.g., \texttt{float} or \texttt{uint8}) and dimension ordering (e.g., channels first or last). To this end, OpenDR has defined and fully follows one standard way of representing data, adopting well-established formats (e.g., those used by PyTorch) when possible. This is enforced during the development by using a number of pre-defined tools for data loading, minimizing the possibility that an algorithm could expect data in a different format.

\noindent \textbf{Trained Models}  OpenDR has selected ONNX as the preferred data format for storing trained models~\cite{onnxruntime}. However, ONNX is often not enough to store metadata related to the models and/or other data structures needed for the function of different algorithms. At the same time, the discrepancies between DL frameworks and the current operators supported by ONNX dictate using native formats in some applications. To this end, OpenDR has defined a data structure for storing models trained with OpenDR toolkit. This data structure relies on existing open standards, while it allows for transparently encapsulating existing formats (e.g., ONNX or native formats used by DL libraries). In this manner, the interoperability with these frameworks is readily ensured, since the users can directly extract the trained models from the OpenDR model format. OpenDR model format is composed of a JSON file,  which contains the necessary metadata, as well as a number of supporting files that are described within this JSON file. 
OpenDR also maintains an individual open repository of open models and data, allowing all algorithms to directly download pretrained models without having the user download them separately.

\noindent \textbf{Distribution}
To maximize the visibility and ease of use of the toolkit, we provide three different ways for installing the toolkit:
\begin{enumerate}
    \item by cloning the GitHub repository,
    \item using the \textit{pip} package manager and PyPI repository,
    \item using \textit{docker}~\cite{boettiger2015introduction}.
\end{enumerate}
The first way provides a fully functional, latest version of the toolkit that can be installed on various platforms. \textit{pip} is a straightforward way to install and experiment with the Python API of the toolkit, while \textit{docker} images are provided to experiment with toolkit functionalities in a pre-configured environment with very little effort, as well as for other containerized applications. Docker images can readily run on all platforms where \textit{docker} is available, while also providing access to GPU acceleration. Furthermore, to ensure the modularity of the toolkit, separate \textit{pip} packages are provided for each submodule of the Python API of the toolkit. For example, if a user needs to use only object detection algorithms, the corresponding package can be used to avoid pulling unnecessary dependencies. Furthermore, \textit{pip} packages are also platform agnostic, allowing for providing both CPU execution and GPU acceleration based on the configuration of the system.  This is ensured by relying on packages that can switch between CPU and GPU acceleration, e.g., PyTorch.

\section{Engine Package}
\label{section:engine}

\begin{figure}
    \centering
    \includegraphics[width=0.99\linewidth]{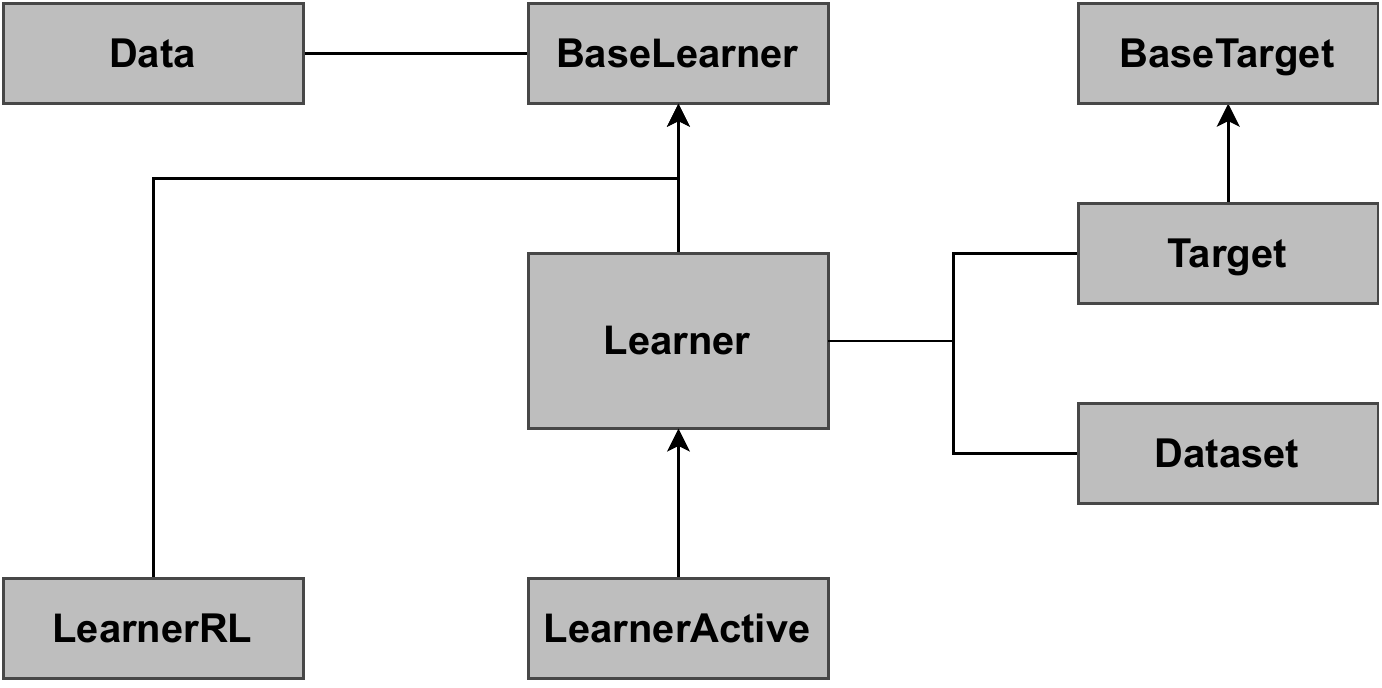}
    \caption{OpenDR abstract class interactions (arrows denote inheritance, lines denote interaction)}
    \label{fig:opendr-classes}
\end{figure}
The engine module contains the necessary abstract classes, as well as various definitions to ensure the homogeneity of OpenDR toolkit, as depicted in Fig.~\ref{fig:opendr-structure} and further detailed in Fig.~\ref{fig:opendr-classes}. More specifically, it contains:
\begin{itemize}
    \item  Abstract class definitions for trainable models (\textit{BaseLearner}, \textit{Learner}, \textit{LearnerActive}).
    \item An abstract class definition for datasets (\textit{Dataset}), as well as a class for supporting native datasets.
    \item An abstract class definition for representing different types of data (\textit{Data}).
    \item Concrete class definitions for representing different types of data  (e.g., \textit{Image}, \textit{Video}, \textit{Timeseries}, \textit{Vector}, \textit{PointCloud}, etc.).
    \item Abstract class definitions for outputs (\textit{Target}), as well as for annotations (\textit{BaseTarget}).
    \item Concrete class definitions for prediction outputs as well as for annotations inheriting the \textit{Target} class (e.g., \textit{Category, Pose, BoundingBox, SpeechCommand, etc.}).
\end{itemize}
This modular structure allows for easily adding new classes to support new functionalities as the development progresses, as well as maintaining the homogeneity of the implementation in an easy way. In the rest of this section, we provide detailed descriptions of the \textit{Learner} classes, along with the \textit{Data}, \textit{Target} and other supporting classes.

\noindent \textbf{Learner Classes}
All classes responsible for implementing any perception algorithm  inherit the abstract \textit{BaseLearner} class to ensure that a common interface is provided for training. These classes should provide methods for:
\begin{enumerate}
    \item  training models (\texttt{fit()}),
    \item evaluating the performance of a trained model (\texttt{eval()}),
    \item performing inference (\texttt{infer()}),
    \item saving/loading a trained model (\texttt{save()}/\texttt{load()}),
    \item optimizing a trained model for inference (\texttt{optimize()}),
    \item downloading pretrained models and supporting data (\texttt{download()}), and
    \item resetting the state of the model, if applicable, (\texttt{reset()}).
\end{enumerate}
\textit{BaseLearner} class provides the interface specifications that are shared among different \textit{Learner} classes. \textit{BaseLearner} also specifies various parameters that can alter the behavior of models during the training and inference process. These parameters, along with any algorithm / model-specific parameters, are carefully documented in OpenDR’s documentation. 
Furthermore,  inference can be either stateless or stateful. For stateful inference algorithms (e.g., recurrent neural networks or tasks like tracking), the \texttt{reset()} method can be used for re-initializing them.  Most perception algorithms are implementing the basic \texttt{Learner} class. This class provides the interface for training, inference, and evaluation methods.  Also, note that \texttt{fit()} and \texttt{eval()} return a standardized dictionary containing statistics regarding the training/evaluation process. 

\noindent \textbf{Data and Dataset Classes} To allow representing different types of data, an abstract \textit{Data} class has been created. This class serves as the basis for the more complicated data types. For data classes, conversion from (using the constructor) and to \textit{NumPy} arrays (using the \texttt{numpy()} method) is supported to make the library compliant with the standard pipelines used by the computer vision and robotics communities. Furthermore, an \texttt{opencv()} function is used for images to easily get the data in an OpenCV-compliant format~\cite{bradski2008learning}. Also, other supporting functions, such as \texttt{open()} for loading the data from files and \texttt{convert()} for performing conversions in a standarized way, ensure that the user will easily provide (and get back) the data in the correct format to (from) the toolkit. \textit{Data} class  serves  as the basis for implementing other data types that are commonly used, such as:
\begin{enumerate}
    \item \textit{Vector} class, for representing one dimensional vectors, 
    \item \textit{Timeseries} class,  containing a series of multi-dimensional measurements,
    \item \textit{Image} and \textit{Video} classes, containing multi-channel images and videos,
    \item \textit{PointCloud} and \textit{PointCloudWithCalibration} class, containing point cloud data (without/with calibration data), and others.
\end{enumerate}

Furthermore, two different types of dataset classes are supported by OpenDR (both inheriting the same abstract \textit{Dataset} class). The first one is the \textit{DatasetIterator}. \textit{DatasetIterator} serves as an abstraction layer over the different types of datasets following PyTorch dataset conventions.
In this way, it provides the opportunity to users to implement different kinds of datasets while providing a uniform interface. Furthermore, the \textit{ExternalDataset}  provides a way for handling well-known external dataset formats (e.g., COCO~\cite{lin2014microsoft}, PascalVOC~\cite{everingham2010pascal}, Imagenet~\cite{russakovsky2015imagenet}, etc.) directly by OpenDR, without requiring any special effort by the users in writing a specific loader.

\noindent \textbf{Target classes} 
The output of various algorithms (predicted targets) is represented using the \textit{Target} class. This class is also used to represent the training annotations, i.e., the targets that are used during the training process.  First, a root \textit{BaseTarget} class has been created to allow for setting the hierarchy of different targets. Classes that inherit \textit{BaseTarget} can be used either as the output of an algorithm or as ground truth annotations, but there is no guarantee that this is always possible, i.e., that both options are always possible. This is in contrast with the \textit{Target} class, which always guarantees that the subclasses can be used for both cases. Therefore, classes that are only used to provide ground truth annotations must inherit from \textit{BaseTarget} instead of Target. Then, to allow representing different types of targets, an abstract \textit{Target} class has been created. This class serves as the basis for more specialized forms of targets. 

Concrete classes inheriting \textit{Target} class are used both to specify training annotations and output data for perception algorithms. Some examples of these classes are the following:
\begin{itemize}
    \item \textit{Category} for simple classification problems,
    \item \textit{BoundingBox} and \textit{BoundingBox3D} for individual bounding boxes,
    \item \textit{Pose} for problems related to human pose estimation and (keypoint-based) body part detection,
    \item \textit{Heatmap} for multi-class segmentation problems or multi-class problems that require heatmap annotations/outputs, and others.
\end{itemize}
Please also note that OpenDR follows the well-established OpenAI Gym interface for interfacing reinforcement learning algorithms~\cite{brockman2016openai}.

\begin{figure}
    \centering
    \includegraphics[width=0.6\linewidth, trim={0.4cm 0.3cm 0.2cm 1.4cm}, clip]{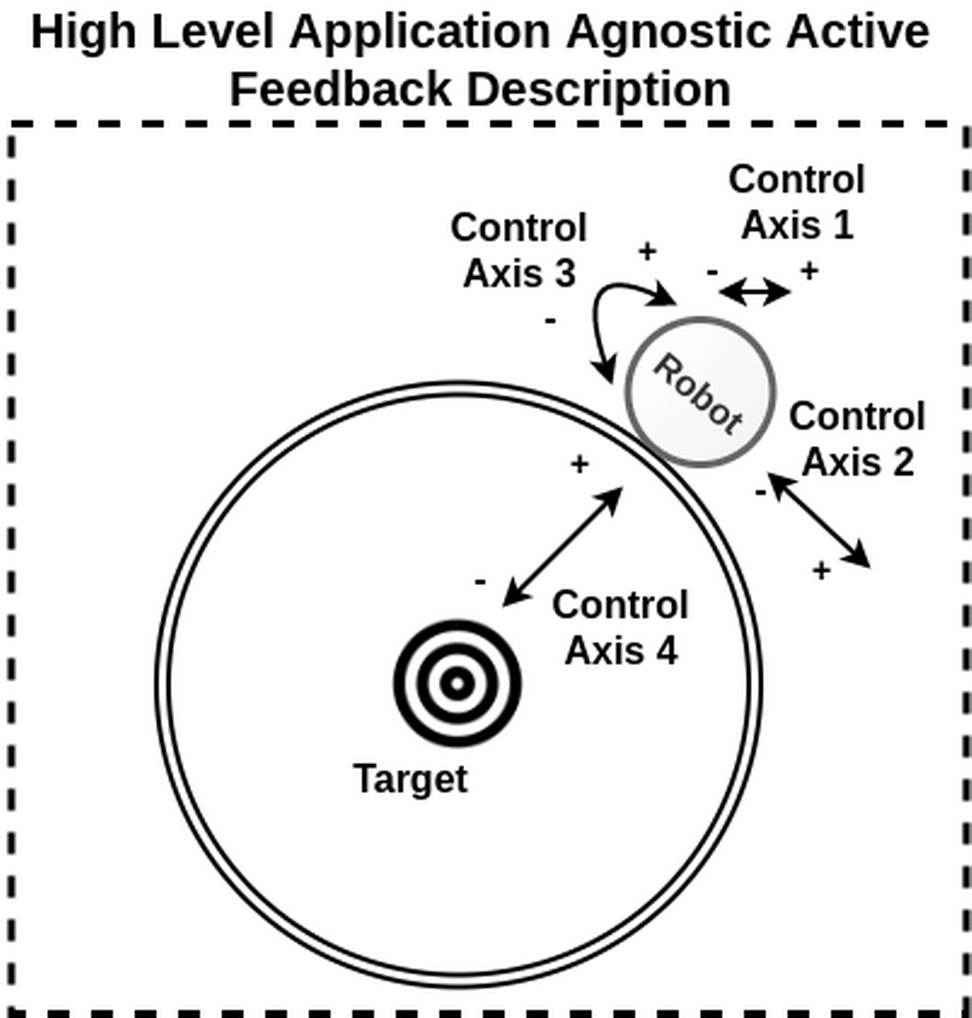}
    \caption{Active perception outputs can be represented in a homogenous way using the proposed application agnostic control specification}
    \label{fig:active}
\end{figure}

\noindent \textbf{Active Perception} A standard for representing active perception outputs has also been specified. To allow algorithms to handle active perception scenarios, each prediction/target class also supports using an additional action field that contains a description of the next action that can be performed to further improve the perception accuracy. All action related classes inherit from an abstract \textit{ActionBase} class. Then, a concrete \textit{Action} class is used to describe actions that can be performed when working with algorithms that support controlling 1-axis only, 2-axes, 3-axes, or 4-axes, in order to improve the perception accuracy. The exact number of axes supported depends on the way each algorithm is trained. For all axes, it is assumed that the robot moves in a sphere and a real value from $-1$ to $1$ is provided for the movement on each axis, as shown in Fig.~\ref{fig:active}. Note the positive sign (+) and negative sign (-) for matching between the positive control interval ($0 \dots 1$) and the negative control interval ($-1 \dots  0$).

\section{USAGE EXAMPLES}
\label{sec:examples}

In this section we provide simple usage examples to demonstrate the ease of use of OpenDR. In the following example we demonstrate how to use the Single Short Detector (SSD)~\cite{liu2016ssd} for person detection:

\footnotesize
\begin{lstlisting}[language=Python]
from opendr.engine.data import Image
from opendr.perception.object_detection_2d import SingleShotDetectorLearner

ssd = SingleShotDetectorLearner(device="cpu")
ssd.download(".", mode="pretrained")
ssd.load("./ssd_default_person")
img = Image.open("example.jpg")
boxes = ssd.infer(img)
\end{lstlisting}
\normalsize
Note that we have the opportunity to select the inference device when initializing the detector, as well as the utility provided for downloading a pretrained person detection model. Then, we use the \textit{Image} class to load an image and perform object detection. We can also use visualization utilities to examine the results as shown below:

\footnotesize
\begin{lstlisting}[language=Python]
from opendr.perception.object_detection_2d import draw_bounding_boxes
draw_bounding_boxes(img.opencv(), boxes, class_names=ssd.classes, show=True)
\end{lstlisting}
\normalsize
It is worth noting that using any tool from OpenDR follows the same pipeline (initialize the model, train/load a pretrained model and then performance inference). For example, for running pose estimation using the OpenPose algorithm~\cite{cao2017realtime}, we can follow the same pipeline:

\footnotesize
\begin{lstlisting}[language=Python]
from opendr.perception.pose_estimation import LightweightOpenPoseLearner as PoseEstimator
from opendr.engine.data import Image

pose_estimator = PoseEstimator(device="cuda")
pose_estimator.download(path=".")
pose_estimator.load("openpose_default")
img = Image.open("example.png")
poses = pose_estimator.infer(img)
print(poses)
\end{lstlisting}
\normalsize
Note that for this tool we selected a GPU accelerator for inference (\texttt{device="cuda"}). Also, OpenDR automatically casts the outputs of the models into strings when passed to the \texttt{print()} function to allow users to easily examine the inference results.

\section{CONCLUSIONS}
\label{sec:conclusions}
In this paper, we presented the design principles of OpenDR, along with the architecture that was developed following these principles. To this end, an abstract interface was created to provide a uniform and homogeneous interface for various robotic tasks, spanning beyond traditional DL cognition and inference, while going towards robotics-oriented perception. OpenDR is actively developed and it is expected that by its third major release it will incorporate active perception for all of the major tools that are implemented. Finally, it is worth mentioning that in less than 2 months after the first release of the toolkit, it has been downloaded/cloned more than 10,000 times, demonstrating its high potential in this rapidly evolving area.

\bibliographystyle{IEEEtran}
\bibliography{bib}

\end{document}